\documentclass[letterpaper]{article} 
\usepackage{aaai25}  
\usepackage{times}  
\usepackage{helvet}  
\usepackage{courier}  
\usepackage[hyphens]{url}  
\usepackage{graphicx} 
\usepackage{natbib}  
\usepackage{caption} 
\usepackage{algorithm}
\usepackage{algorithmic}
\usepackage{amsmath}
\usepackage{booktabs}       
\usepackage{multirow}
\usepackage{xcolor}
\usepackage{newfloat}
\usepackage{listings}
\DeclareCaptionStyle{ruled}{labelfont=normalfont,labelsep=colon,strut=off} 
\floatstyle{ruled}
\newfloat{listing}{tb}{lst}{}
\floatname{listing}{Listing}
\title{RAP-SR: RestorAtion Prior Enhancement in Diffusion Models \\ for Realistic Image Super-Resolution}

\author {
Jiangang Wang\textsuperscript{\rm 1,\rm 2}, Qingnan Fan\textsuperscript{\rm 2\dag}, Jinwei Chen\textsuperscript{\rm 2}, Hong Gu\textsuperscript{\rm 2}, Feng Huang\textsuperscript{\rm 3}, Wenqi Ren\textsuperscript{\rm 1\dag}\\
}
\affiliations {
    \textsuperscript{\rm 1}School of Cyber Science and Technology, Shenzhen Campus of Sun Yat-sen University\\
    \textsuperscript{\rm 2}vivo Mobile Communication Co. Ltd\\
    \textsuperscript{\rm 3}School of Mechanical Engineering and Automation, Fuzhou University\\
    wangjg33@mail2.sysu.edu.cn, fqnchina@gmail.com, renwq3@mail.sysu.edu.cn\\
   {\small Project Page: \url{https://github.com/W-JG/RAP-SR}}
}

\usepackage{bibentry}

\begin{document}

\maketitle

\begin{abstract}
Benefiting from their powerful generative capabilities, pretrained diffusion models have garnered significant attention for real-world image super-resolution (Real-SR).
Existing diffusion-based SR approaches typically utilize semantic information from degraded images and restoration prompts to activate prior for producing realistic high-resolution images.
However, general-purpose pretrained diffusion models, not designed for restoration tasks, often have suboptimal prior, and manually defined prompts may fail to fully exploit the generated potential.
To address these limitations, we introduce RAP-SR, a novel restoration prior enhancement approach in pretrained diffusion models for Real-SR. 
First, we develop the High-Fidelity Aesthetic Image Dataset (HFAID), curated through a Quality-Driven Aesthetic Image Selection Pipeline (QDAISP). Our dataset not only surpasses existing ones in fidelity but also excels in aesthetic quality.
Second, we propose the Restoration Priors Enhancement Framework, which includes Restoration Priors Refinement (RPR) and Restoration-Oriented Prompt Optimization (ROPO) modules.
RPR refines the restoration prior using the HFAID, while ROPO optimizes the unique restoration identifier, improving the quality of the resulting images.
RAP-SR effectively bridges the gap between general-purpose models and the demands of Real-SR by enhancing restoration prior.
Leveraging the plug-and-play nature of RAP-SR, our approach can be seamlessly integrated into existing diffusion-based SR methods, boosting their performance. Extensive experiments demonstrate its broad applicability and state-of-the-art results.
\textit{Codes and datasets will be available upon acceptance.}
  \vspace{-0.5cm}
\end{abstract}

%
\begin{figure}[t]
    \centering
    \includegraphics[width=0.9\linewidth]{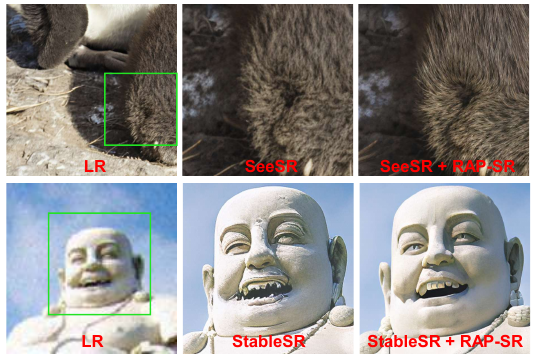}
      \caption{
Visual Comparison: RAP-SR enhances the restoration prior of pretrained diffusion models. Our proposed RAP-SR method can be seamlessly integrated into diffusion-based SR methods, generating more realistic details and textures without the need for fine-tuning the original model.}
  \label{fig:figure_1}
  \vspace{-0.7cm}
\end{figure}

\renewcommand{\thefootnote}{}
\footnotetext{\dag \ Corresponding author.}
\footnote{This work was completed during an internship at vivo.}

\section{Introduction}
Image super-resolution (SR) is a fundamental task in computer vision, aiming to reconstruct high-resolution (HR) images from low-resolution (LR) inputs, with broad applications in mobile photography~\cite{chen2019camera}, autonomous driving~\cite{li2023azimuth}, and robotics~\cite{wang2021real}. SR remains a highly ill-posed problem due to the complexity and variability of degradation models in real-world scenarios. Early SR solutions focus on improving fidelity~\cite{SRCNN,kim2016deeply,haris2018deep} by employing pixel-level losses such as \(\ell\)$_1$ and MSE, often resulting in over-smoothed details~\cite{liang2022details}. 
Advanced architectures~\cite{Swinir} have improved performance, but issues like artifacts and poor visual quality remain when applied to real-world scenarios.
To address this, Real-SR methods aim to reproduce realistic details by optimizing both fidelity and perceptual quality, often employing generative adversarial networks (GANs)~\cite{GAN}. However, GAN-based approaches suffer from problems such as model collapse and difficult training~\cite{li2022best}. 

Recently, diffusion models~\cite{DDPM} have gained prominence in image generation, leading to the development of large-scale pretrained text-to-image (T2I) diffusion models, such as Stable Diffusion (SD). 
For a wide range of natural images, diffusion prior is more effective than GAN-based prior. Additionally, these models have shown significant potential in various downstream low-level vision tasks, including image editing~\cite{sdedit,avrahami2022blended}, image restoration~\cite{lugmayr2022repaint,chung2022come}, and image-to-image translation~\cite{song2020score,saharia2022palette}. Methods such as StableSR~\cite{stablesr}, PASD~\cite{pasd}, DiffBIR~\cite{diffbir}, SUPIR~\cite{supir}, and SeeSR~\cite{seesr} leverage pretrained T2I models to tackle the Real-SR problem by capturing semantic structures from LR images and using handcrafted restoration prompts to activate restoration prior for generating realistic HR images.

However, employing pretrained diffusion models for Real-SR tasks presents two primary challenges: the inadequacy of restoration prior and inaccuracies in prompt activation.
General pretrained diffusion models are not inherently designed for restoration tasks. While these models have strong prior knowledge and can generate images of varying quality, their limited restoration prior hinders their ability to produce high-quality, rich-detail images~\cite{emu,pixart}. 
Moreover, previous diffusion-based SR methods often rely on manually crafted restoration prompts\footnote{\textit{e.g.}, SeeSR positive prompt: ``clean, high-resolution, 8k''; negative prompt: ``dotted, noise, blur, lowres, smooth''. SUPIR positive prompt: ``cinematic, high contrast, highly detailed, 32k, ultra HD, extreme meticulous detailing, \textit{etc}''; SUPIR negative prompt: ``painting, oil painting, illustration, drawing, art, sketch, \textit{etc}''.} to activate restoration prior. Natural language often fails to accurately describe image quality under multiple degradations, leading to incorrectly activated restoration prior.

To address these limitations, we propose \textbf{RAP-SR}, a novel restoration prior enhancement approach in pretrained diffusion models for Real-SR.
Firstly, we develop the \textit{High-Fidelity Aesthetic Image Dataset (HFAID)} using a Quality-Driven Aesthetic Image Selection Pipeline (QDAISP). 
Although large-scale datasets are available, the images within these datasets often suffer from poor and inconsistent quality due to varied purposes. 
We propose a meticulous four-stage image selection process (QDAISP) facilitated by a large-scale multi-modality model to filter images by evaluating both the image quality and aesthetic attributes. As a result, HFAID consists of 5,000 high-fidelity and aesthetic images from a pool of 1 million, surpassing the quality of all existing datasets tailored for image restoration.
This dataset serves as the foundation for enhancing the prior of pretrained models, enabling the transition from low-quality image generation to high-quality output production.
Secondly, we establish a \textit{Restoration Prior Enhancement framework}, including restoration prior refinement (RPR) and restoration-oriented prompt optimization (ROPO). RPR refines the restoration prior by fine-tuning the model using HFAID. ROPO optimizes specific identifiers during the prior refinement phase. 
The method combines unique identifiers with the image's semantic caption, strengthening the association between prompt and image quality, and enabling accurate activation of the restoration prior.
As a result, our framework effectively strengthens the restoration prior, bridging the gap between general-purpose models and Real-SR tasks. 

RAP-SR's plug-and-play design allows it to be seamlessly integrated with existing diffusion-based SR methods, such as StableSR~\cite{stablesr}, DiffBIR~\cite{diffbir}, and SeeSR~\cite{seesr}, improving both visual quality and objective metrics.
Overall, our contribution is summarized as follows:
\begin{itemize}
\item We collected the High-Fidelity Aesthetic Image Dataset (HFAID), which surpasses existing datasets not only in fidelity but also in aesthetic quality, effectively enhancing the priors of diffusion models.

\item We proposed a Prior Restoration Enhancement framework, which includes the Restoration Prior Refinement (RPR) and Restoration-Oriented Prompt Optimization (ROPO) modules, designed to improve and accurately activate the model's restoration priors.

\item Our method can be seamlessly integrated to improve existing diffusion-based SR methods. Extensive experiments demonstrate its broad applicability and excellent performance.
\end{itemize}


\begin{figure*}[t]
  \centering
  \includegraphics[width=1.0\linewidth]{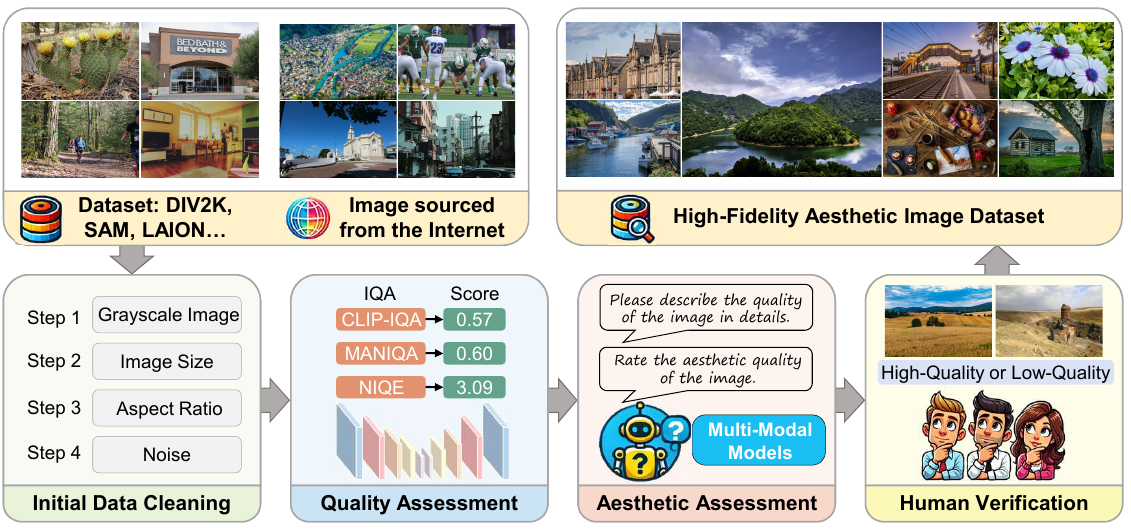}
  \caption{Quality-Driven Aesthetic Image Selection Pipeline. This process is divided into four stages. Unlike previous methods that focus solely on image quality, our approach incorporates the multi-modality model to evaluate both image quality and aesthetic performance. Ultimately, we meticulously select 5,000 ultra-high-quality images from the initial pool of one million images to create the High-Fidelity Aesthetic Image Dataset.
  }
  \label{fig:dataset_pipeline}
  \vspace{-0.5cm}
\end{figure*}

\section{Related Work}

\subsection{Real-world Image Super Resolution}
Deep learning has emerged as the predominant approach for SR tasks, with the pioneering work of SRCNN~\cite{SRCNN} utilizing deep neural networks for SR.
Subsequent methods that incorporate residual connections and attention mechanisms~\cite{Swinir,kim2016deeply,haris2018deep} often aim to minimize fidelity loss through pixel-level supervised loss functions. However, this approach typically results in overly smoothed details~\cite{liang2022details}.
Recent research has shifted towards addressing Real-SR challenges, which involve complex and unknown degradation processes.
Some researchers propose collecting real-world LR and HR paired data to train networks~\cite{RealSR, DrealSR}, although this approach can be costly. Alternatively, other methods focus on synthesizing realistic data pairs for training. Notably, BSRGAN~\cite{Bsrgan} and RealESRGAN~\cite{realesrgan} offer efficient degradation pipelines for Real-SR.
Given the generative capabilities of GAN-based methods~\cite{GAN}, they have become dominant in Real-SR tasks. Adversarial training improves the perceptual quality of images. However, GANs face limitations such as training instability and model collapse, which often result in unnatural artifacts~\cite{li2022best}. Consequently, recent research has begun exploring advanced generative models, such as diffusion models, which are capable of generating high-quality, detailed images.

\subsection{Diffusion Model}
Diffusion models utilize Markov chains to transform latent variables into complex data distributions, as exemplified by DDPM~\cite{DDPM} and its accelerated variant, DDIM~\cite{ddim}. The Latent Diffusion Model (LDM)\cite{ldm} achieves impressive results with reduced computational costs. These advancements enable large-scale pretrained text-to-image (T2I) models like Stable Diffusion (SD) and ImgGen. ControlNet~\cite{controlnet} allows for external control over the generation process, while EMU~\cite{emu} enhances aesthetic quality through fine-tuning. InstructPix2Pix~\cite{instructpix2pix} refines T2I models using editing instructions. Diffusion models excel in various image generation tasks, including restoration~\cite{lugmayr2022repaint,chung2022come}, editing, and colorization~\cite{song2020score,saharia2022palette}.

\subsection{Diffusion-Based Super-Resolution}
Diffusion-based SR methods can be categorized into three main types. The first type adjusts the inverse process of pretrained diffusion models using gradient descent~\cite{wang2022zero,kawar2022denoising,fei2023generative}.
These methods do not require retraining but assume a predefined image degradation model, limiting their applicability in Real-SR scenarios. The second type involves training the diffusion model on paired data~\cite{ldm,shang2024resdiff,resshift}, but the restoration quality is heavily dependent on the quantity of training data, constraining their potential to achieve exceptional results. 
The third type leverages the robust generative prior of large-scale pretrained diffusion models by introducing adapters for control~\cite{seesr,supir,pasd}. By utilizing LR images as control information, pretrained diffusion models can produce high-quality results, making this approach the mainstream for diffusion model-based SR methods. StableSR~\cite{stablesr} balances fidelity and perceptual quality by incorporating a time-aware encoder and feature warping. DiffBIR~\cite{diffbir} employs SwinIR~\cite{Swinir} for initial degradation removal, enhancing details with a diffusion model. PASD~\cite{pasd} utilizes semantic models like ResNet~\cite{resnet} to extract information from LR images, thereby bolstering the generative capability of T2I models. SeeSR~\cite{seesr} improves T2I model generation using tags and additional conditions. CoSeR~\cite{coser} augments T2I model generation by providing reference images from LR inputs. SUPIR~\cite{supir} achieves superior outcomes through the use of a larger diffusion model, coupled with a robust language model. These methods guide T2I diffusion models in generating high-quality HR images by extracting additional semantic information from degraded LR images. However, they often overlook the restoration prior inherent in pretrained diffusion models, which are crucial for image reconstruction tasks.
\section{Methodology}

\begin{figure*}[t]
  \centering
  \includegraphics[width=1.0\linewidth]{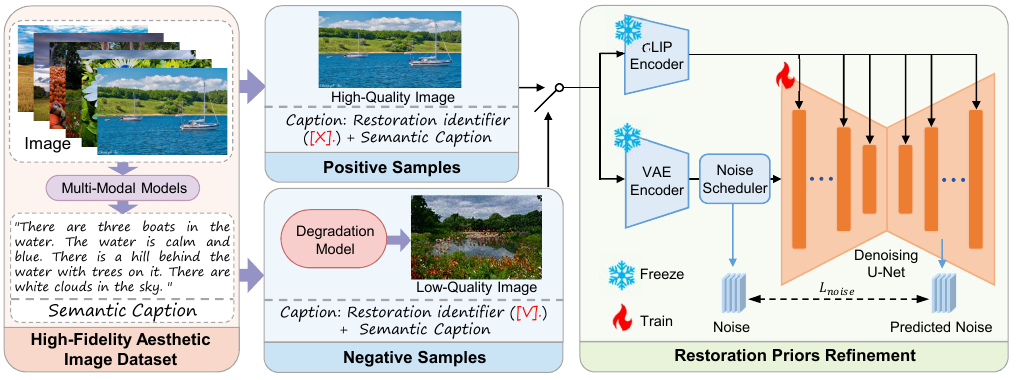}
  \caption{
Restoration Priors Enhancement Framework: This framework includes Restoration Priors Refinement (RPR) and Restoration-Oriented Prompt Optimization (ROPO). ROPO optimizes the restoration prompt by constructing both positive and negative samples. For negative samples, a degradation model generates low-quality images and then combines the unique restoration identifier with the image's semantic caption to create training data. Through the subsequent RPR process, the model enhances its restoration prior, learning to associate image quality with the restoration identifier.
  }
  \label{fig:frame_work}
  \vspace{-0.5cm}
\end{figure*}


\subsection{High-Fidelity Aesthetic Image Dataset}
\subsubsection{Observation and Motivation}
Previous research~\cite{emu,pixart} indicates that training large-scale T2I diffusion models involves multiple phases. The initial phase focuses on aligning text and images, where the diffusion model establishes a mapping between the two by leveraging billions of text-image pairs. The subsequent phase, known as quality-tuning, aims to enhance image quality. Once text-image alignment is achieved, the pretrained model can be fine-tuned using a small dataset tailored to specific task domains. In the Real-SR task, it is particularly important to enhance the model's restoration prior through quality-tuning.

During the quality-tuning phase, the quality of the dataset is crucial to the effectiveness of model training~\cite{emu}. An ideal dataset should contain high-quality, detail-rich images with informative captions. However, existing datasets like LAION-5B~\cite{laion-5b} suffer from poor image quality, incomplete captions, and misalignment between images and text. Widely adopted datasets such as SAM~\cite{SAM}, COCO~\cite{COCO}, and ImageNet~\cite{ImageNet} also have low quality and inadequate labeling. Although datasets such as DIV2K~\cite{div2k}, Flickr2K~\cite{flick2k}, and LSDIR~\cite{lsdir} provide relatively high quality, they still do not meet ultra-high-quality standards due to quality inconsistencies in their datasets.

\subsubsection{Quality-Driven Aesthetic Image Selection Pipeline}
To address these issues, we curate a high-fidelity aesthetic image dataset (HFAID) by selecting 5,000 ultra-high-quality images from an initial pool of 1 million images. To effectively filter the image data, we design a quality-driven aesthetic image selection pipeline that considers both image quality and aesthetic performance. As shown in Figure~\ref{fig:dataset_pipeline}, the selection process is divided into four stages: Initial Data Cleaning, Quality Assessment, Aesthetic Assessment, and Human Verification.

Initially, we collect approximately 1 million images from existing datasets and publicly available online sources. Initial data cleaning is performed, including checks for grayscale images, verification of image size and aspect ratio, and the use of Laplacian variance~\cite{lsdir} to detect image noise.
In the second stage, to accurately assess image quality, we employ state-of-the-art no-reference image quality assessment metrics. The currently available no-reference image quality metrics (e.g., CLIP-IQA, MANIQA, MUSIQ, NIQE, BRISQUE, etc.) each focus on different aspects. 
Our goal is to select extremely high-quality data that meets human aesthetic standards. Therefore, the choice of metrics is crucial. We first select 200 images with the best and worst performance under each metric from the LSDIR dataset and conduct a 10-person user evaluation, ultimately selecting MANIQA~\cite{maniqa}, CLIP-IQA~\cite{clipiqa}, and NIQE~\cite{niqe} as the core evaluation metrics.

Previous image restoration datasets primarily focus on image quality and detail richness but lack exploration of aesthetic evaluation. The aesthetic quality of images is equally crucial for image generation tasks. Studies have shown that multi-modality models have surpassed traditional models in the field of image understanding~\cite{llava}. In the third phase, we use existing multi-modality models for aesthetic evaluation, utilizing the mPLUG-Owl2~\cite{mplugowl2} model to query images and obtain precise aesthetic evaluation metrics.
In the final phase, we conduct human verification to accurately assess the quality of each image, ensuring that each image is evaluated by at least two people. We provide a detailed explanation of the selection process in the supplementary material.

\begin{figure}[t]
  \centering
  \includegraphics[width=0.7\linewidth]{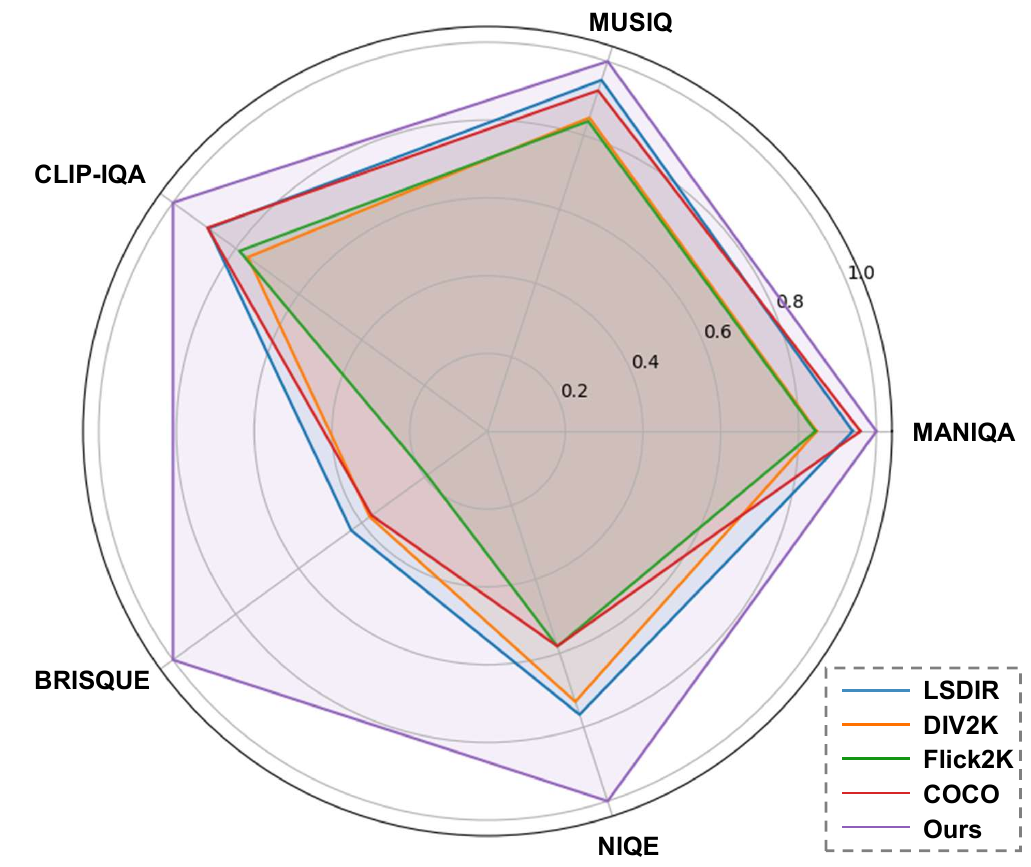}
  \caption{
  Comparison of No-Reference Metrics Across Different Datasets. Our proposed dataset significantly outperforms existing datasets across all evaluation metrics.
  }
  \label{fig:dataset_compare}
  \vspace{-0.7cm}
\end{figure}

\subsubsection{Comparison With Other Datasets }
As illustrated in figure~\ref{fig:dataset_pipeline}, our dataset excels in both image quality and detail richness.  
We evaluate the quality of our dataset using five no-reference image quality assessment metrics: MANIQA, MUSIQ, CLIP-IQA, BRISQUE, and NIQE. Figure ~\ref{fig:dataset_compare} presents the results, showing that our dataset significantly outperforms others across all metrics. Additional examples are provided in the supplementary material.

\subsection{Restoration Priors Enhancement Framework}
\subsubsection{Restoration Priors Refinement}
\label{Refinement}
High-quality image captions are also crucial for training diffusion models. To generate text labels with high information density, we utilize the advanced vision-language model Florence-2~\cite{florence}, a large-scale end-to-end multi-modality model. Leveraging its image understanding capabilities, we produce high-quality text labels. The quality of our labels surpasses that of existing text-image datasets~\cite{laion-5b}, with further details provided in the supplementary materials.

Since the pre-trained diffusion model has already completed text-image alignment, we can achieve quality-tuning with high-quality image-text data in a short period, thereby enhancing the model's ability to restoration prior knowledge. Using a smaller batch size of 40, the model converges within 3,000 steps, significantly reducing training time. Additionally, we find that both the quantity and quality of data significantly influence the tuning effect, which will be discussed in detail in section \ref{sec:ablation}.

\subsubsection{Restoration-Oriented Prompt Optimization}
Previous diffusion-based SR models often rely on manually designed restoration prompts to activate restoration prior, frequently resulting in defocused images and artifacts that degrade the overall quality of image restoration. DreamBooth~\cite{dreambooth} introduces a novel method for fine-tuning pre-trained diffusion models by associating a unique identifier with a specific object using a small number of images, enabling the generation of realistic images that accurately represent the object. While DreamBooth focuses on binding specific object concepts to pretrained diffusion models, we extend its application to restoration prompt optimization for image quality. We develop a restoration-oriented prompt optimization method to precisely activate the model's restoration prior. The method is shown in figure~\ref{fig:frame_work}.

Given the challenges of fully expressing image quality through natural language, especially under conditions of multiple degradations, we create new identifiers to represent various levels of image quality. During the restoration prior enhancement phase in diffusion models, we redefine the categories of high-quality and degraded low-quality image data to better align with the needs of models.

For the high-quality category, we utilize high-quality image data and combine positive restoration identifiers with the semantic caption of the images to form positive samples. These positive samples serve as a benchmark for the model to understand what constitutes high-quality imagery. In contrast, for the low-quality category, we first generate degraded low-quality image data using a sophisticated image degradation model. We then combine negative restoration identifiers with the semantic captions of the original images to create negative samples. By merging these unique identifiers with their semantic captions, we effectively link image quality with image semantics, making it easier for the model to distinguish between different quality levels.

During the training process, we perform a random sampling of positive and negative samples according to a predetermined ratio $r$. This is crucial for fine-tuning the pre-trained diffusion model, enabling it to learn how to associate high-quality and low-quality images with their corresponding identifiers. As a result, the model becomes adept at accurately activating the restoration prior, which significantly enhances its ability to generate high-quality images from degraded inputs. This method ensures a more reliable and consistent image restoration process across various scenarios.
The pseudo-code of our restoration-oriented prompt optimization algorithm is summarized as Algorithm~\ref{alg:algorithm}.

During inference, we adopt a classifier-free guidance strategy, which enables the diffusion model to generate higher-quality images using negative prompts without additional training. At each inference step, we calculate the positive prompt $c_{pos}$ and negative prompt $c_{neg}$ and mix these predictions to obtain the final output:
\begin{gather}
    \hat{\epsilon} =  \epsilon_{\theta}(z_{lr}^{t}, t, c_{pos}, x_{lr}), \\
    {\hat{\epsilon}_{\text{neg}}} = \epsilon_{\theta}(z_{lr}^{t}, t, c_{neg}, x_{lr}),  \\
    \tilde{\epsilon} = \hat{\epsilon} + \lambda_{s}(\hat{\epsilon} - {\hat{\epsilon}_{\text{neg}}} ). 
\end{gather}
where \(\lambda_s\) is the guidance scale and \(z_{lr}^t\) represents the noise potential of the low-resolution image. In practice, We employ unique identifiers defined during our training as positive prompts \(c_{pos}\) and negative prompts \(c_{neg}\) to generate higher-quality images.

\begin{algorithm}[t]
\caption{Restoration-oriented Prompt Optimization Algorithm}
\begin{algorithmic}[1]
\REQUIRE Ultra-high-quality Dataset of image-text pairs $S = \{(x_i, c_i)\}_{i=1}^N$, diffusion model $f_\theta$, degradation model $d_\mu$, number of timesteps $T$, noise schedule $\{\beta_t\}_{t=1}^T$, positive identifier $c_p$, negative identifier $c_n$, positive ratio $r$, learning rate $\eta$
\STATE Initialize model parameters $\theta$ from a pre-trained model

\FOR{each $(x_i, c_i)$ in $S$}
    \STATE Sample a random value $u \sim \text{Uniform}(0, 1)$
    \IF{$u < r$}
        \STATE Append positive identifier: \\
        $c_i \gets \text{concatenate}(c_p, c_i)$
    \ELSE
        \STATE Append negative identifier: \\
        $c_i \gets \text{concatenate}(c_n, c_i)$
        \STATE Degrade the image: $x_i \gets d_\mu(x_i)$
    \ENDIF
    \STATE Sample timestep $t \sim \text{Uniform}(1, T)$
    \STATE Sample noise $\epsilon \sim \mathcal{N}(0, I)$
    \STATE Compute noisy image: $x_t \gets \sqrt{\bar{\alpha}_t} \cdot x_i + \sqrt{1 - \bar{\alpha}_t} \cdot \epsilon$, where $\bar{\alpha}_t = \prod_{s=1}^{t} (1 - \beta_s)$
    \STATE Predict noise: $\hat{\epsilon}_\theta \gets f_\theta(x_t, t, c_i)$ 
    \STATE Compute loss: $L \gets \|\epsilon - \hat{\epsilon}_\theta\|^2$
    \STATE Update model parameters: $\theta \gets \theta - \eta \nabla_\theta L$
\ENDFOR
\end{algorithmic}
\label{alg:algorithm}

\end{algorithm}

\begin{table*}[h]
\centering
\small
\begin{tabular}{c|c|cc|cc|cc}
\toprule
Datasets                 & Metric  & StableSR        & Stable + RAP-SR  & DiffBIR         & DiffBIR + RAP-SR & SeeSR            & SeeSR + RAP-SR   \\
\midrule
\multirow{8}{*}{DIV2K}   & PSNR $\uparrow$    & 23.28         & \textbf{23.85} & 23.66         & \textbf{23.83} & \textbf{23.70} & 23.59          \\
                         & SSIM $\uparrow$    & 0.5733          & \textbf{0.5808}  & 0.5651          & \textbf{0.5694}  & \textbf{0.6052}  & 0.5897           \\
                         & LPIPS $\downarrow$  & \textbf{0.3118} & 0.3542           & \textbf{0.3516} & 0.3536           & \textbf{0.3168}  & 0.3501           \\
                         & MANIQA $\uparrow$  & \textbf{0.6193} & 0.6017           & 0.6211          & \textbf{0.6255}  & 0.6246           & \textbf{0.6271}  \\
                         & MUSIQ $\uparrow$   & 65.85        & \textbf{66.32} & 65.77         & \textbf{66.90} & 68.66          & \textbf{68.86} \\
                         & CLIPIQA $\uparrow$ & 0.6771          & \textbf{0.7517}  & 0.6693          & \textbf{0.6928}  & 0.6936           & \textbf{0.7254}  \\
                         & BRISQUE $\downarrow$ & 15.62       & \textbf{11.84} & 14.66         & \textbf{8.69}  & 20.41          & \textbf{16.08} \\
\midrule
\multirow{8}{*}{RealSR}  & PSNR $\uparrow$    & 24.60         & \textbf{25.03} & 24.81         & \textbf{25.06} & \textbf{25.00} & 24.85         \\
                         & SSIM $\uparrow$    & 0.7045          & \textbf{0.7081}  & 0.6595          & \textbf{0.6663}  & \textbf{0.7187}  & 0.7003           \\
                         & LPIPS $\downarrow$   & \textbf{0.3096} & 0.3457           & 0.3608          & \textbf{0.3564}  & \textbf{0.3096}  & 0.3388           \\
                         & MANIQA $\uparrow$  & 0.6252          & \textbf{0.6259}  & 0.6238          & \textbf{0.6349}  & 0.6456           & \textbf{0.6512}  \\
                         & MUSIQ $\uparrow$   & 66.00         & \textbf{67.39}  & 64.93         & \textbf{67.46} & 69.93          & \textbf{70.0644} \\
                         & CLIPIQA $\uparrow$ & 0.6315          & \textbf{0.6956}  & 0.6448          & \textbf{0.6911}  & 0.6553           & \textbf{0.7217}  \\
                         & BRISQUE $\downarrow$ & 19.51          & \textbf{15.39} & 18.98        & \textbf{13.36} & 29.06         & \textbf{19.01} \\
\midrule
\multirow{8}{*}{DrealSR} & PSNR $\uparrow$    & 27.99       & \textbf{28.3019} & 26.82         & \textbf{27.15} & 27.98         & \textbf{28.01} \\
                         & SSIM $\uparrow$    & 0.7504          & \textbf{0.7566}  & 0.6633          & \textbf{0.6858}  & \textbf{0.7719}  & 0.7600             \\
                         & LPIPS $\downarrow$   & \textbf{0.3275} & 0.3644           & 0.4497          & \textbf{0.4365}  & \textbf{0.3243}  & 0.3542           \\
                         & MANIQA $\uparrow$  & 0.5620           & \textbf{0.5667}  & 0.5924          & \textbf{0.6173}  & 0.5941           & \textbf{0.6161}  \\
                         & MUSIQ $\uparrow$   & 59.03       & \textbf{60.70} & 60.85        & \textbf{63.7761} & 64.95          & \textbf{65.08} \\
                         & CLIPIQA $\uparrow$ & 0.6385          & \textbf{0.6752}  & 0.6369          & \textbf{0.6729}  & 0.6757           & \textbf{0.7075}  \\
                         & BRISQUE $\downarrow$ & 21.39         & \textbf{17.36} & 22.60         & \textbf{15.60} & 32.12          & \textbf{23.21} \\
\bottomrule
\end{tabular}
\caption{Quantitative comparison. The best results of each group are highlighted in \textbf{bold}. $\uparrow$ and $\downarrow$ mean that the larger or smaller score is better, respectively.}
\label{tab:fine_tuned_result}
  \vspace{-0.5cm}
\end{table*}

\section{Experiments}

\begin{figure*}[t]
    \centering
    \includegraphics[width=\linewidth]{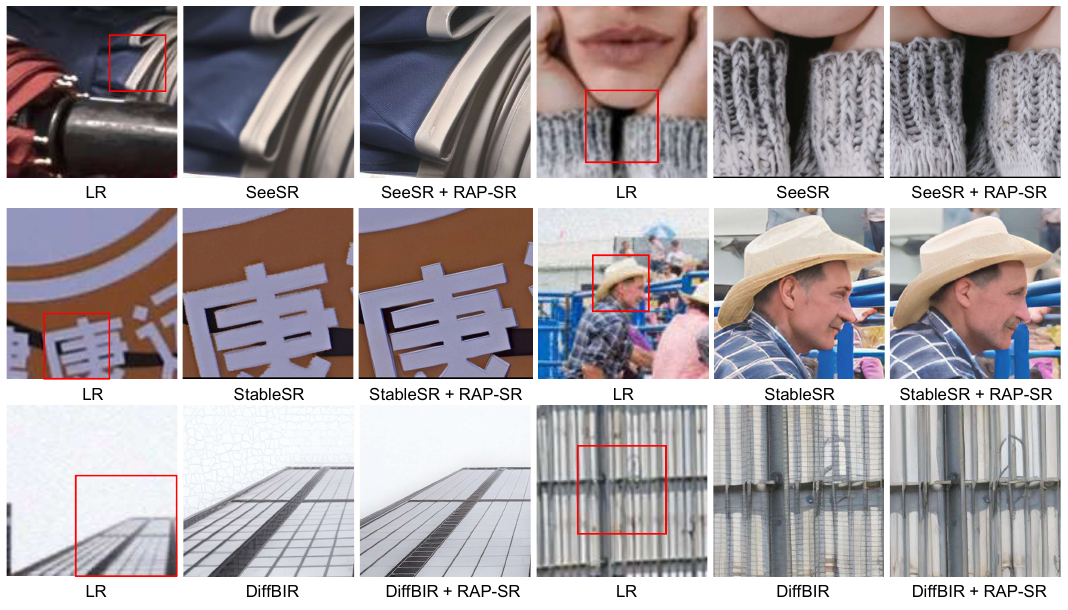}
    \caption{Qualitative comparisons on real-world test datasets. RAP-SR obtains the best visual performance.}
    \label{fig:fine_tuned_result}
  \vspace{-0.2cm}
\end{figure*}

\subsubsection{Enhancing Restoration Prior For Pretrained Diffusion Model.} We conduct the proposed restoration prior enhancement experiments using Stable Diffusion 2.1\footnote{https://huggingface.co/stabilityai/stable-diffusion-2-1} and our HFAID dataset. As described in Section~\ref{Refinement}, we use the Florence-2~\cite{florence} model to generate highly informative text labels. During the enhancement process, the images are resized to 512 pixels on the longer side and center-cropped. Low-quality images are synthesized through the RealESRAGN~\cite{realesrgan} pipeline.
In the optimization process, we employ the AdamW~\cite{adamw} optimizer with a learning rate of 5e-5 and train the model using two NVIDIA L40 GPUs with a batch size of 40 for 3,000 iterations. The ratio $r$ of high-quality (HQ) to low-quality (LQ) images is set at $0.8$, where '[X]' denotes positive identifier and '[V]' denotes negative identifier. To maintain the consistency of CFG~\cite{cfg}, there is a $5\%$ chance of leaving the text labels empty.

\subsubsection{Evaluation Setting.} To evaluate the proposed restoration prior enhancement method, we test three methods: SeeSR, DiffBIR, and StableSR. The results are obtained for SeeSR + RAP-SR, DiffBIR + RAP-SR, and StableSR + RAP-SR. All these methods use Stable Diffusion as the pre-trained model, with their original models replaced by our RAP model. \textbf{It is important to note that we do not perform any fine-tuning on these replaced models.}
During the evaluation, we conduct comprehensive tests on synthetic data from the DIV2K 
 val dataset and real-world datasets RealSR~\cite{RealSR} and DrealSR~\cite{DrealSR}. In the tests, the resolution of high-resolution images is set to 512 × 512, while low-resolution images are cropped to 128 × 128.

\subsubsection{Evaluation Metrics.} To better align with human perception, we use seven evaluation metrics: PSNR, SSIM~\cite{ssim}, LPIPS~\cite{lpips}, MANIQA~\cite{maniqa}, MUSIQ~\cite{musiq}, BRISQUE~\cite{brisque} and CLIPIQA~\cite{clipscore}. PSNR and SSIM measure pixel-level differences, while LPIPS assesses perceptual distances. MANIQA, MUSIQ, BRISQUE, and CLIPIQA are no-reference image quality metrics. Previous studies have shown that reference-based metrics have a weaker correlation with human perception of image quality in real-world scenarios~\cite{supir,stablesr}. A discussion of the test metrics is detailed in the supplementary material.

\subsection{Comparison with state-of-the-arts}
\subsubsection{Quantitative Comparison}
Table~\ref{tab:fine_tuned_result} provides quantitative comparisons on three synthetic and real-world datasets. We have the following observations. Firstly, the method we proposed has achieved great improvements in almost all no-reference metrics such as MANIQA, MUSIQ, CLIPIQA, and BRISQUE on all three data sets. This shows that our method significantly improves the image generation capabilities of the original method and can generate richer details. Secondly, for reference metrics such as PSNR, SSIM, and LPIPS, our method only improves the original performance on some data sets. This is primarily because the DM-based method generates more realistic details, which impact these metrics. Overall, our PAR-SR achieves better no-reference metric scores while maintaining competitive full-reference metric scores.
\subsubsection{Qualitative Comparison}
Figure~\ref{fig:fine_tuned_result} shows visual examples from synthetic and real-world datasets. The visual results are consistent with the quantitative findings: our model significantly improves perceptual quality, produces more realistic textures, and enhances the overall realism of images (e.g., sweaters and landscapes). Furthermore, our method significantly reduces issues such as blurring and artifacts (e.g., in windows and skies). In summary, RAP-SR enables diffusion-based super-resolution methods to more accurately reconstruct image details in real-world scenes. More visual results are provided in the supplementary material.

\begin{table}[t]
\resizebox{\linewidth}{!}{
\begin{tabular}{c|c|cc}
\toprule
                              & Configurations  & MANIQA $\uparrow$ & CLIPIQA $\uparrow$ \\

\midrule
\multirow{3}{*}{Dataset size} & 1000             & 0.5681 & 0.639   \\
                              & 3000            & 0.5727 & 0.6423  \\
                              & 8000            & 0.5926 & 0.6866  \\
\midrule
\multirow{3}{*}{Prompt}       & w/o prompt           & 0.5806 & 0.6308  \\
                              & w/o negative prompt  & 0.6063 & 0.6864  \\
                              & w/o positive prompt  & 0.6107 & 0.6629  \\
\midrule
Ours                      & Default         & \textbf{0.6161} & \textbf{0.7075}  \\
\bottomrule
\end{tabular}}
\caption{Ablation study. Test on the DrealSR dataset.}
\label{tab:Ablation}
  \vspace{-0.3cm}
\end{table}

\subsection{Ablation Study}
\label{sec:ablation}
Due to the superior generative capabilities of SeeSR~\cite{seesr}, all ablation experiments for our model are conducted using the SeeSR model.

\subsubsection{Effect of Different Dataset Sizes }
We conducted random split tests on the proposed HFAID dataset with varying sizes, as shown in Table~\ref{tab:Ablation}. When the dataset size is small, there is a significant decline in no-reference metrics. When the dataset size is expanded to 8,000 images, the model begins to converge, and both reference and no-reference metrics show a decline compared to the default 5,000-image dataset, resulting in further performance degradation.
\subsubsection{Effect of Different Prompts}
We test our fine-tuned T2I model using various restoration prompts to validate their effects. The prompts included: positive prompt only, negative prompt only, and no prompt. The results are shown in Table~\ref{tab:Ablation}. Firstly, we observe a significant drop in no-reference perceptual quality metrics when no prompts are used, underscoring the crucial role of restoration prompts. Additionally, negative prompts prove more beneficial for image generation compared to positive prompts.
When we use both positive and negative prompts, all metrics achieve optimal results.

\section{Conclusion}
This paper introduces RAP-SR, a novel approach that enhances restoration priors in pretrained diffusion models for real-world image super-resolution (Real-SR) tasks. We develop the High-Fidelity Aesthetic Image Dataset (HFAID) through a Quality-Driven Aesthetic Image Selection Pipeline (QDAISP), surpassing existing datasets in fidelity and aesthetic quality. The Restoration Priors Enhancement Framework, including Restoration Priors Refinement (RPR) and Restoration-Oriented Prompt Optimization (ROPO), refines priors and optimizes restoration identifiers. RAP-SR seamlessly integrates into diffusion-based SR methods, significantly boosting performance. Extensive experiments demonstrate its broad applicability and state-of-the-art results.

\bibliography{aaai25}

\appendix
\section*{Appendix}
\section{Details of the HFAID}
In this section, we provide a detailed comparison of our proposed High-fidelity Aesthetic Image Dataset (HFAID) with existing datasets, highlighting differences in visual quality, quantitative metrics, and caption accuracy.
\subsection{Quantitative Comparisons}
To accurately assess the effectiveness of our proposed  HFAID Dataset, we conduct a comparison of no-reference metrics across different datasets, as shown in Table~\ref{tab:dataset}. In our selection pipeline, we use CLIPIQA~\cite{clipiqa}, MANIQA~\cite{maniqa}, and NIQE~\cite{niqe} as the primary evaluation metrics. To avoid data bias, we also employ MUSIQ~\cite{musiq} and BRISQUE~\cite{brisque} as additional testing metrics. The results indicate that our dataset achieves the best outcomes across all metrics. Compared to the previously high-quality restoration dataset LSDIR~\cite{lsdir}, our dataset shows significant improvements across multiple no-reference metrics: The NIQE metric improves by 23\%, the BRISQUE improves by 56\% and the CLIPIQA improves by 11\%. This demonstrates the advantages of our proposed dataset.
\subsection{Qualitative Comparisons}
\begin{figure*}[t]
  \centering
  \includegraphics[width=1.0\linewidth]{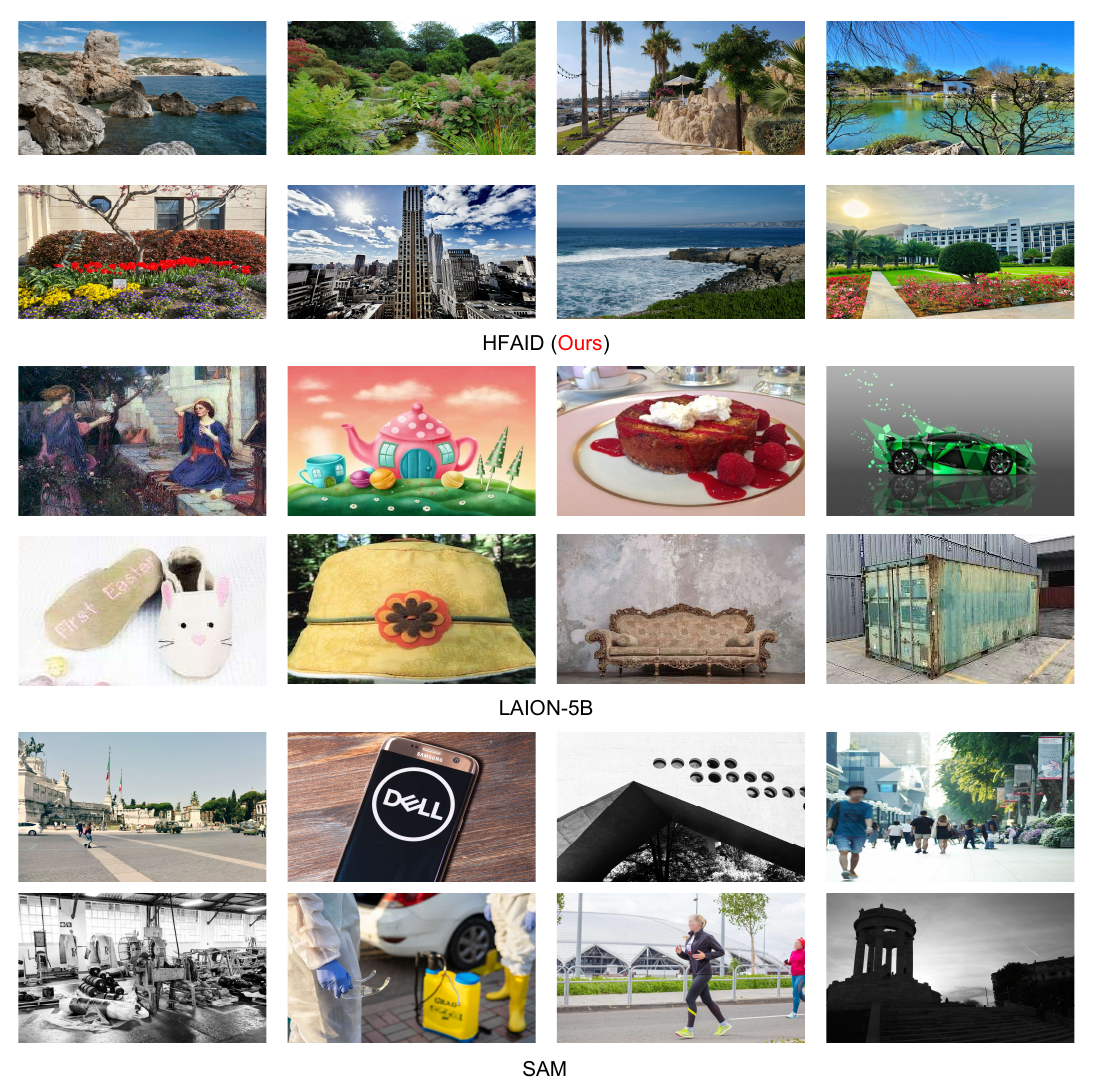}
  \caption{Qualitative Comparisons on different datasets. Our HFAID dataset is compared with the LAION-5B and SAM datasets. The results demonstrate that the HFAID dataset excels in image quality and aesthetic performance. In contrast, the LAION-5B dataset shows lower image quality, while the SAM dataset has even poorer quality, with facial features of individuals intentionally obscured for privacy reasons.}
 \label{fig:dataset_compare_2}
\end{figure*}

In Figure~\ref{fig:dataset_compare_2}, we present the visual results of our dataset. To ensure a fair comparison, we randomly sample eight images from the dataset for comparison. Compared to existing datasets, the HAFID dataset excels in both image quality and aesthetic performance. Specifically, the HAFID dataset demonstrates higher image quality and better alignment with human aesthetic preferences than the commonly used LAION-5B~\cite{laion-5b} dataset in text-to-image tasks. The SAM~\cite{SAM} dataset applies blurring to all faces to protect privacy, but this significantly impacts the dataset's usefulness in model training.

\subsection{Comparison of Image Captions}
\begin{figure}[t]
    \centering
    \includegraphics[width=0.9\linewidth]{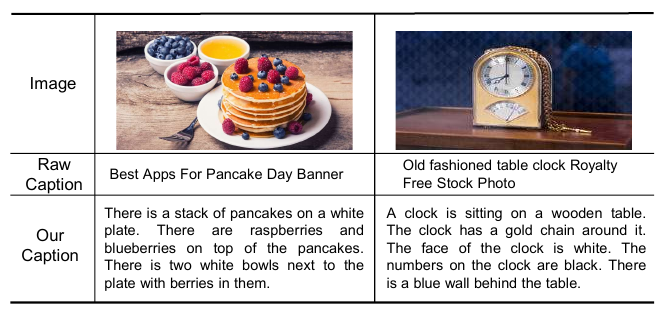}
      \caption{Comparison of Different Image Captions. In existing text-image datasets, such as LAION, the caption quality is generally low and prone to errors. In contrast, we utilize advanced vision-language models to generate more refined captions, thereby providing richer data for model training.}
  \label{fig:caption}
\end{figure}

In the training process of Diffusion models, the quality of text labels is just as important as image quality. To better describe the images, we use advanced vision-language models to generate high-density captions. As shown in Figure~\ref{fig:caption}, the existing LAION dataset contains numerous errors in its captions, which are often brief and fail to fully describe the image's content. In contrast, the captions we generate show significant improvements in both descriptive quality and information density, enabling our model to perform more effectively in quality control during generation.

\begin{table}[t]
\resizebox{\linewidth}{!}{
\begin{tabular}{c|ccccc}
\toprule
        & MANIQA$\uparrow$ & MUSIQ$\uparrow$    & CLIPIQA$\uparrow$  & BRISQUE$\downarrow$  & NIQE $\downarrow$    \\
        \midrule

DIV2K   & 0.6041      & 64.17 & 0.5729 & 15.14 & 3.092   \\
Filck2K & 0.6017     & 63.43 & 0.5933   & 29.54 & 3.886 \\
COCO    & 0.6844     & 69.73 & 0.6696 & 15.40 & 3.891 \\
LSIDR   & 0.6702     & 71.86 & 0.6675 & 13.08 & 2.950 \\ 
\midrule
Ours    & \textbf{0.7136}    & \textbf{75.69}  & \textbf{0.7524} & \textbf{5.671} & \textbf{2.260} \\
\bottomrule
\end{tabular}}
\caption{Quantitative comparisons across different datasets. The results indicate that our dataset achieves the best performance across all metrics.}
\label{tab:dataset}
  \vspace{-0.3cm}
\end{table}

\section{Details of the QDAISP}
In the second phase of the Quality-Driven Aesthetic Image Selection Pipeline (QDAISP), our goal is to accurately assess image quality in alignment with human aesthetic preferences. The core of this process lies in identifying image quality assessment metrics that best match human evaluation standards. To achieve this, we conduct a detailed user study.

\begin{figure}[t]
    \centering
    \includegraphics[width=0.5\linewidth]{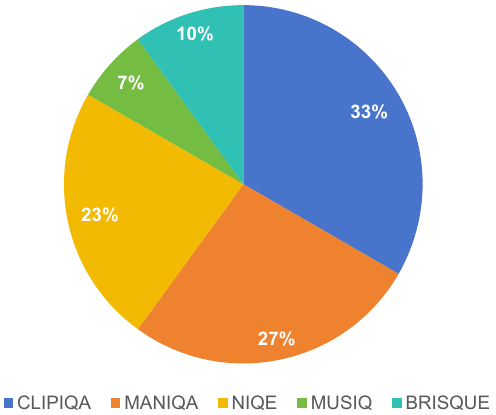}
      \caption{User Study Results. The voting results of this study are based on feedback from 10 volunteers. The image quality assessment metrics that best align with human quality preferences are identified by evaluating the performance of the images and the metrics.}
  \label{fig:user_study}
\end{figure}

\begin{figure*}[t]
  \centering
  \includegraphics[width=0.8\linewidth]{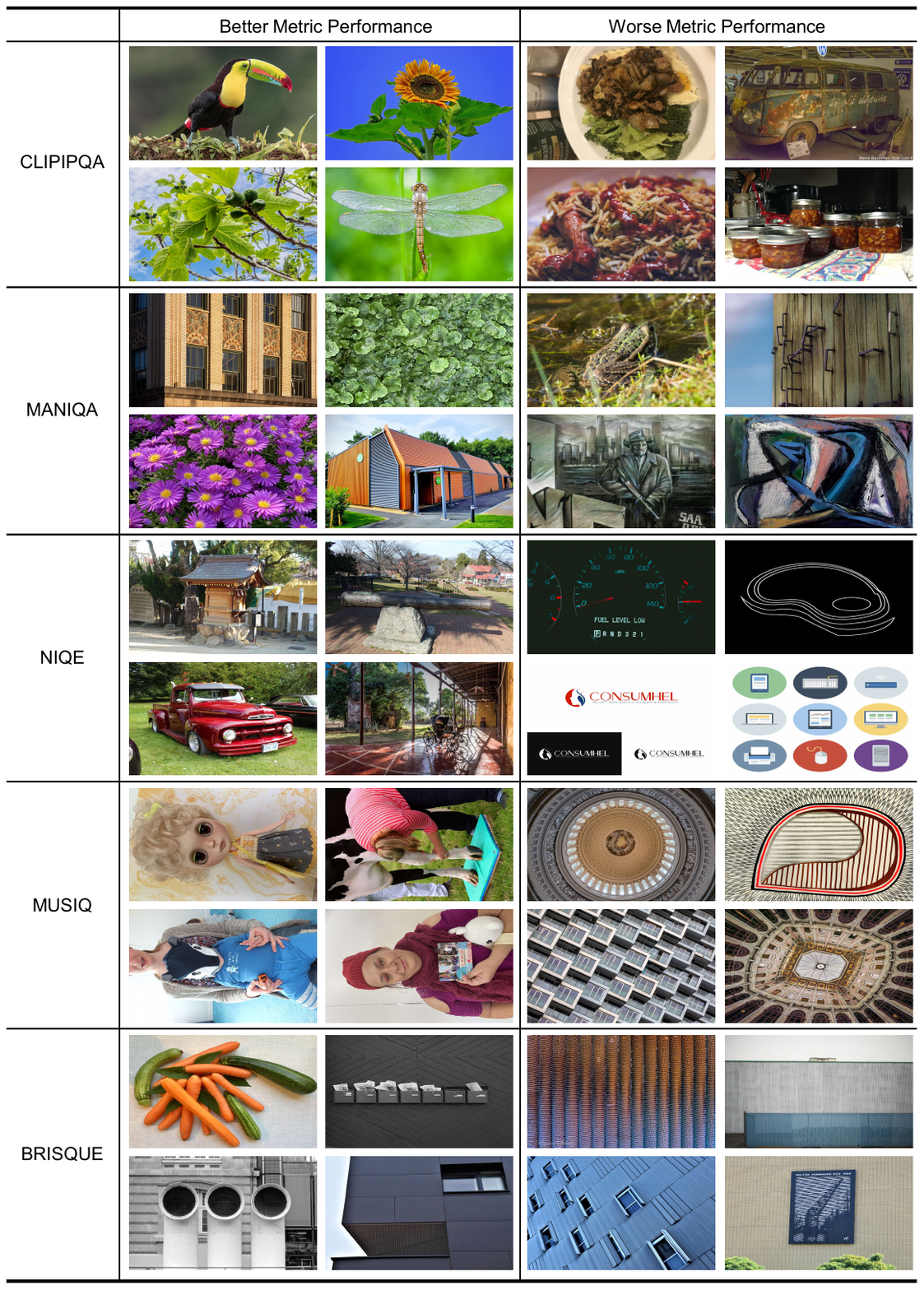}
  \caption{Image quality assessment and its corresponding images used in the user study. The study results indicate that CLIPIQA, MANIQA, and NIQE more accurately reflect human aesthetic preferences and excel at distinguishing between high-quality and low-quality images. In contrast, metrics like MUSIQ and BRISQUE demonstrate poor separability between different image qualities and diverge from human aesthetic preferences.}
  \label{fig:metric_compare}
\end{figure*}

First, we test the LSDIR~\cite{lsdir} dataset using commonly used no-reference metrics, such as CLIPIQA, MANIQA, MUSIQ, NIQE, and BRISQUE. We then select 200 images from both the best and worst-performing results for each metric for analysis, as shown in Figure~\ref{fig:metric_compare}. Next, we organize a group of 10 researchers to rate these metrics to identify those that most accurately reflect human aesthetic preferences. After voting, CLIPIQA, NIQE, and MANIQA are selected as the key metrics for evaluating image quality. The voting results are presented in Figure~\ref{fig:user_study}. It is important to note that we do not claim BRISQUE and MUSIQ are useless in image quality assessment; rather, we emphasize that metrics like CLIPIQA more closely align with human aesthetic preferences.

As shown in Figure~\ref{fig:metric_compare}, images with better metric performance in CLIPIQA and MANIQA exhibit richer details, brighter visuals, prominent subjects, and a sense of aesthetic appeal. In contrast, images with worse metric performance display noticeable color casts and poorer image quality. NIQE effectively identifies synthetic images, particularly in worse metric performance. Furthermore, we find that BRISQUE and MUSIQ correlate less with human aesthetic preferences.

\begin{figure}[ht]
    \centering
    \includegraphics[width=0.9\linewidth]{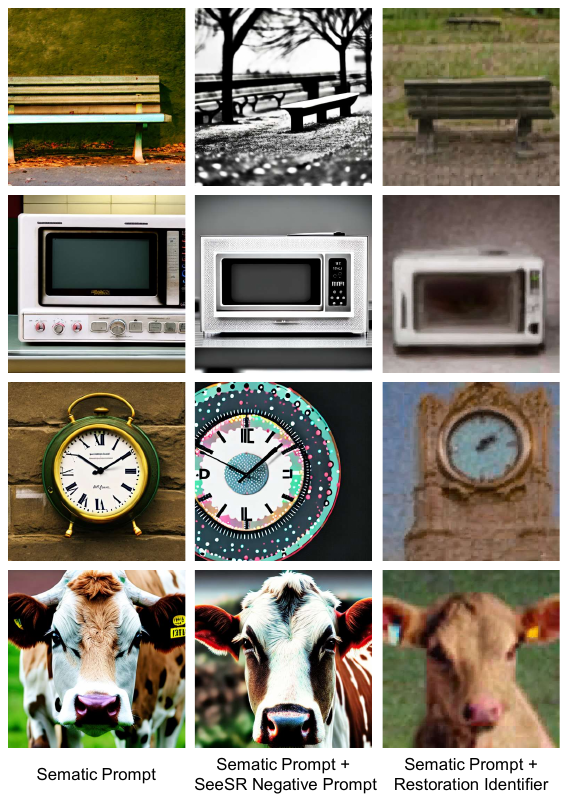}
      \caption{Comparison of the Effectiveness of Our ROPO Method on T2I Tasks. When we apply different restoration prompts (such as SeeSR's negative prompts: ``dotted, noise, blur, lowres, smooth'') and our method (Restoration Negative Identifier), ROPO generates more realistic degradation effects than SeeSR prompts. Consequently, this allows the diffusion model to produce more realistic results when utilizing CFG.}
  \label{fig:generate_result}
\end{figure}


\begin{table}[t]
\small
\resizebox{\linewidth}{!}{
\begin{tabular}{c|ccccc|c}
\toprule
Metric  & LAION  & Flick2K & SAM    & DIV2K  & LSDIR  & HFAID           \\
\midrule
MANIQA  & 0.6434 & 0.6247  & 0.6275 & 0.6475 & 0.6333 & \textbf{0.6512} \\
CLIPIQA  & 0.6258 & 0.6683  & 0.6841 & 0.6806 & 0.6923 & \textbf{0.7217} \\
\bottomrule
\end{tabular}}
\caption{Effectiveness of HFAID. Our proposed HFAID significantly improves the results.}
\label{tab:dataset_ablation}
\end{table}

\section{Ablation Study}
This section provides additional ablation experiments on High-fidelity Aesthetic Image Dataset (HFAID) and Restoration-Oriented Prompt Optimization (ROPO).

\subsection{Effectiveness of HFAID}
We train RAP-SR on various datasets to demonstrate the superiority of our proposed HFAID. The datasets used include DIV2K~\cite{div2k}, LAION-5B, SAM, Flick2K~\cite{flick2k}, LSDIR, and our HFAID dataset. Each dataset is trained for the same number of epochs to ensure fairness. We use SeeSR as the base model, train RAP-SR with different datasets, and test it on the RealSR dataset. As shown in Table~\ref{tab:dataset_ablation}, the HFAID dataset significantly improves the model's performance across multiple metrics compared to other datasets. This underscores the significant advantage of our dataset in enhancing result quality.

\subsection{Effectiveness of Restoration-Oriented Prompt Optimization}
To validate the effectiveness of our proposed Restoration-Oriented Prompt Optimization (ROPO) strategy, we conduct tests on text-to-image tasks using default semantic prompts, SeeSR's negative prompts, and our Restoration Identifier. As shown in Figure~\ref{fig:generate_result}, SeeSR's negative prompts produce simple degradation effects like grayscale images, while our Restoration Identifier generates more realistic degradations. These effects are often hard to describe precisely with language, but our optimized approach enables the Diffusion model to correlate prompts with realistic degradations strongly.

During the inference phase, by using the Classifiers-Free Guidance (CFG)~\cite{cfg} strategy, which combines positive and negative prompts, the model generates more realistic results. Positive prompts guide the model to produce images that match the target description, while negative prompts help steer the output away from undesired features. By optimizing the Restoration Identifier, the model effectively avoids real degradation features, leading to generate more realistic images.

\section{Misalignment Between Human Perception and Image Quality Assessment Metrics}
\begin{figure}[t]
    \centering
    \includegraphics[width=0.7\linewidth]{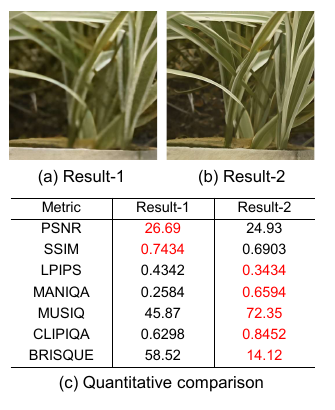}
      \caption{Comparison of No-Reference and Full-Reference Metrics. In two different results (Result-1 and Result-2), although Result-1 performs better on full-reference metrics such as PSNR and SSIM, it fails to deliver highly realistic outcomes compared to Result-2. Therefore, in Real-SR tasks, no-reference metrics are more valuable than full-reference metrics.}
  \label{fig:metric_error}
\end{figure}

In Figure~\ref{fig:metric_error}, we present an illustrative example that highlights the differences between image quality assessment metrics and human perception. We compare two different results and evaluate them using multiple metrics. Quantitative evaluation shows that Result-2 scores lower than Result-1 on full-reference metrics, such as PSNR and SSIM, but higher on no-reference metrics, such as CLIPIQA and MANIQA. However, visual assessment reveals that Result-2 produces a more realistic effect than Result-1, effectively reducing smoothness and blurring. This suggests that no-reference image quality assessment metrics, such as MANIQA, MUSIQ, and CLIPIQA, align more closely with human visual perception trends. This further underscores the importance of no-reference image quality assessment metrics in real-world super-resolution tasks.

\section{Additional Visual Results from RAP-SR}
\begin{figure*}[t]
  \centering
  \includegraphics[width=1.0\linewidth]{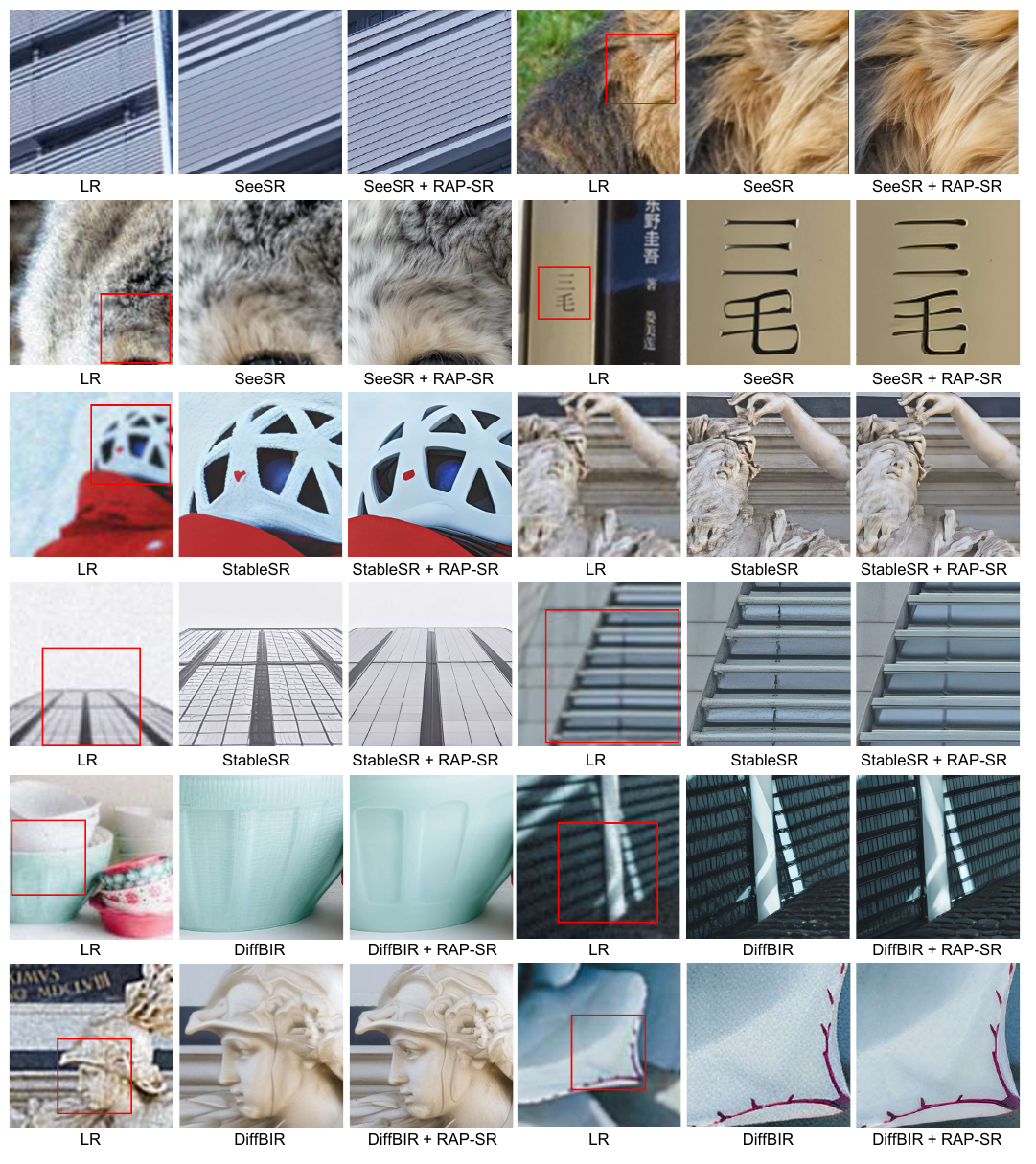}
  \caption{Qualitative comparisons on different test datasets. RAP-SR obtains the best visual performance. Please zoom in for a detailed view.}
  \label{fig:result_compare}
\end{figure*}

In Figure~\ref{fig:result_compare}, we present additional visual results to demonstrate the superior applicability and performance of RAP-SR.

\end{document}